\pdfoutput=1

\documentclass[11pt]{article}

\usepackage[preprint]{acl}
\usepackage{times}
\usepackage{latexsym}
\usepackage[T1]{fontenc}
\usepackage[utf8]{inputenc}
\usepackage{microtype}
\usepackage{inconsolata}
\usepackage{graphicx}
\usepackage{amsmath}
\usepackage{booktabs}
\usepackage{multicol}
\usepackage{xcolor}
\usepackage{tabularray}
\usepackage{arydshln}
\usepackage{hyperref}
\usepackage[most]{tcolorbox}
\usepackage{setspace}
\usepackage{rotating}
\usepackage{lipsum}
\usepackage{soul}
\usepackage{float}

\usepackage{tikz}
\usepackage{tikz-dependency}
\usetikzlibrary{arrows, positioning, calc}

\usepackage{outlines}

\newcommand{\all}[1]{{\color{gray} [all]: #1}}
\newcommand{\assignto}[1]{{\color{red} [#1]}}
\newif\ifshowacks
\showackstrue

\newcommand{\udnewscrawl}[1]{\textsc{UD-NewsCrawl}}
\newcommand{\upd}[1]{AnonUniversityA}
\newcommand{\koln}[1]{AnonUniversityB}
\newcommand{\huggingface}{\raisebox{-1.5pt}{\includegraphics[height=1.05em]{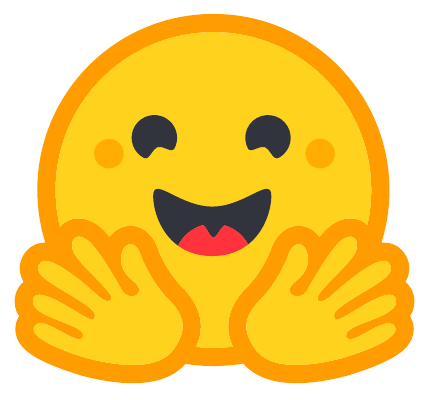}}}

\newcommand*\samethanks[1][\value{footnote}]{\footnotemark[#1]}
\tikzset{deprel/.style={->,>=stealth,thick}}

\makeatletter
\DeclareRobustCommand{\PHP}{%
  \begingroup
  \leavevmode\,\vphantom{P}%
  \dimen\z@=.5\fontcharht\font`P\relax
  \dimen\tw@=0.33333\dimen\z@
  \ooalign{%
    \raisebox{\dimexpr\dimen\z@+2\dimen\tw@-0.4pt}{\rule{\fontcharwd\font`P}{0.4pt}}\cr
    \raisebox{\dimexpr\dimen\z@+\dimen\tw@-0.2pt}{\rule{\fontcharwd\font`P}{0.4pt}}\cr
    P\cr
  }%
  \,\endgroup
}
\makeatother

\title{
    The \udnewscrawl{} Treebank: Reflections and Challenges from a Large-scale Tagalog Syntactic Annotation Project
}

\author{
  \textbf{Angelina A. Aquino}\textsuperscript{2,4}\thanks{Equal contributions.}\hspace{0.5em}
  \textbf{Lester James V. Miranda}\textsuperscript{1}\samethanks{}\hspace{0.5em}
  \textbf{Elsie Marie T. Or}\textsuperscript{3}\samethanks{}\hspace{0.5em}
  \\
  \textsuperscript{1}Allen Institute for AI
  \textsuperscript{2}Charles Darwin University\\
  \textsuperscript{3}Department of Linguistics, University of the Philippines Diliman\\
  \textsuperscript{4}Electrical and Electronics Engineering Institute, University of the Philippines Diliman
  \\
  \texttt{angelina.aquino@cdu.edu.au,ljvmiranda@gmail.com,etor@up.edu.ph}
  \\\\
  {\small\huggingface{}~~\textbf{Dataset}\hspace{0.5em}\href{https://hf.co/datasets/UD-Filipino/UD_Tagalog-NewsCrawl}{\texttt{hf.co/datasets/UD-Filipino/UD\_Tagalog-NewsCrawl}}}
}

\begin{document}
\maketitle

\begin{abstract}
  This paper presents \udnewscrawl{}, the largest Tagalog treebank to date, containing 15.6k trees manually annotated according to the Universal Dependencies framework.
  We detail our treebank development process, including data collection, pre-processing, manual annotation, and quality assurance procedures.
  We provide baseline evaluations using multiple transformer-based models to assess the performance of state-of-the-art dependency parsers on Tagalog.
  We also highlight challenges in the syntactic analysis of Tagalog given its distinctive grammatical properties, and discuss its implications for the annotation of this treebank.
  We anticipate that \udnewscrawl{} and our baseline model implementations will serve as valuable resources for advancing computational linguistics research in underrepresented languages like Tagalog.
\end{abstract}

\section{Introduction}

The Philippine archipelago is home to over 170 languages.
Among these, Tagalog is the most well-known and widely spoken, as the native language of the country's capital region and a ``\textit{de facto} national working language'' \citep{ethnologue} of education, governance, trade, and many other domains of communication.
The native speakers of Tagalog reside mostly in the southwestern regions of Luzon, the Philippines' largest island, but Tagalog has also been adopted as a second language in many other parts of the country \citep{Himmelmann2005}.
Roughly 25 percent of Filipinos (out of a national population of over 100 million) identify as being of Tagalog ethnicity, while 40 percent of households in the country use it as one of their home languages \citep{psa-2024-census}.
Large Tagalog-speaking populations have also been recorded in many other countries, including over 1.3 million Tagalog speakers in the United States and over 700,000 speakers each in Canada and Saudi Arabia \citep{ethnologue}.

Despite the widespread use of Tagalog, large-scale NLP resources for the language remain sparse \citep{joshi-etal-2020-state,cruz-cheng-2022-improving,miranda-2023-developing}.
At present, only two Tagalog treebanks, annotated using the Universal Dependencies (UD) framework \citep{de-marneffe-etal-2021-universal, nivre-etal-2020-universal}---a standardized annotation scheme for linguistic data---are available for public use: UD-Ugnayan \citep{aquino-de-leon-2020-parsing} and UD-TRG \citep[\textit{Tagalog Reference Grammar},][]{Samson2018TRG}.
Together, these treebanks contain just 149 sentences, which are dwarfed by treebanks of high-resource languages and are insufficient for training robust dependency parsers.
Furthermore, Tagalog treebanks lag behind those of its Southeast Asian neighbors in terms of size, such as Indonesian (7.6k sentences), Vietnamese (3.4k), and Thai (1.0k).\footnote{\url{https://universaldependencies.org/}}
This underscores a need for more comprehensive computational resources to support the development of Tagalog language technologies.

In this work, we present \udnewscrawl{}, the largest Tagalog treebank to date, consisting of 15,619 sentences and manually annotated in accordance with the UD framework.
The treebank is composed of text sourced from the Leipzig Tagalog NewsCrawl corpus.\footnote{\url{https://corpora.uni-leipzig.de/en?corpusId=tgl\_newscrawl\_2011\#tgl}}
The text was annotated by native speakers with domain knowledge in linguistics, and the annotations were verified through manual inspection and semi-automated quality control.
Furthermore, we introduce transformer-based parsers on \udnewscrawl{} which serve as baselines for further research into Tagalog dependency parsing.
The treebank and baseline models are available in this \href{https://huggingface.co/collections/UD-Filipino/universal-dependencies-for-tagalog-67573d625baa5036fd59b317}{HuggingFace collection}.

\section{Related Work}
\label{sec:relatedwork}

\paragraph{Universal Dependencies (UD) and Tagalog treebanks.}
The Universal Dependencies framework aims to provide a consistent annotation schema for parts-of-speech (POS) tagging, morphological features, and dependency relations across languages.
UD falls under the class of dependency grammars, and follows a principle of content word primacy \citep{muischnek-etal-2016-estonian, dirix-etal-2017-universal}, i.e. syntactic relations in UD primarily hold between content words, and syntactic structures are typically headed by content words, with function words being dependencies thereof.

As of writing, there are two Tagalog treebanks which apply this framework: UD-Ugnayan \citep{aquino-de-leon-2020-parsing} and UD-TRG \citep{Samson2018TRG}.
The former contains 94 sentences from educational fiction and non-fiction text drawn from the Philippine Department of Education's Learning Resource Portal, while the latter contains 55 sentences from Tagalog grammar books \citep{Schachter2023TagalogRG,deVos2010Essential}.
However, due to their size, the UD dataset guidelines recommend these treebanks to be treated as test data (and to use 10-fold cross-validation for evaluation if one wishes to train on the dataset), and this limits scalable development of NLP models.

\paragraph{Tagalog dependency parsing.}
Several works have used UD-Ugnayan and UD-TRG for automated dependency parsing.
For example, \citet{aquino-de-leon-2020-parsing} showed that even a small treebank like UD-Ugnayan can be used to develop dependency parsers, trained using UDPipe \citep{straka-etal-2016-udpipe} and Stanza \citep{qi-etal-2020-stanza}, that are competitive with cross-lingual or multilingual parsers.
They have also shown that decent tokenization and tagging performance can be achieved using alternative language resources and data augmentation \citep{aquino-de-leon-2022-zero}.
On the other hand, \citet{miranda-2023-calamancy} trained a dependency parser by combining both treebanks and using the spaCy framework \citep{spacy} based on pre-trained RoBERTa embeddings \citep{conneau-etal-2020-unsupervised} .
Due to the absence of a canonical train, development, and test split, prior works resort to different evaluation paradigms such as k-fold cross-validation that may not be directly comparable, hampering effective benchmarking and consistent assessment of parser performance across studies.

\section{Background}

\subsection{The Tagalog language}
\label{sec:linguistics}

\begin{figure*}[t]
  \centering
  \begin{tabular}{lllllll}
    \textit{Nag-bigay}   & \textit{\textbf{ang}} & \textit{\textbf{lalaki}} & \textit{ng}           & \textit{bulaklak}          & \textit{sa}           & \textit{babae}          \\
    \textsc{av.prf}-give & \textsc{nom}          & man                      & \textsc{gen}          & flower                     & \textsc{loc}          & woman                   \\ \midrule
    \textit{B<in>igay}   & \textit{ng}           & \textit{lalaki}          & \textit{\textbf{ang}} & \textit{\textbf{bulaklak}} & \textit{sa}           & \textit{babae}          \\
    \textsc{<pv.prf>}give & \textsc{gen}         & man                      & \textsc{nom}          & flower                     & \textsc{loc}          & woman                   \\ \midrule
    \textit{B<in>igy-an} & \textit{ng}           & \textit{lalaki}          & \textit{ng}           & \textit{bulaklak}          & \textit{\textbf{ang}} & \textit{\textbf{babae}} \\
    \textsc{<prf>}give-\textsc{lv}               & \textsc{gen}             & man                   & \textsc{gen}               & flower                & \textsc{nom}             & woman                   \\
  \end{tabular}
  \caption{
    Example sentences illustrating features of Tagalog voice marking under a symmetrical voice analysis: each sentence has a different \textbf{subject} (preceded by the \textbf{\textit{ang}} marker) and a different verbal affix denoting the thematic role (\textsc{av} = agent, \textsc{pv} = patient, \textsc{lv} = locative) of the subject, but all three examples are pragmatically equivalent to the English sentence ``The man gave flowers to the woman.''
  }
  \label{fig:examples}
\end{figure*}

Tagalog, an Austronesian language belonging to the Western Malayo-Polynesian branch of the language family \citep{blust1991}, occupies a prominent position within the field of linguistics, having been documented since the Spanish occupation of the Philippines in the 16th century, and playing a part in the development of the American structuralist movement the beginning of the 20th century \cite{javier2022tagalog}.
It has become the most well-known representative of the so-called ``Philippine-type languages''---a designation for a group of Austronesian languages that share a distinctive grammatical ``voice marking'' system \citep{Himmelmann2005,reid2005tagalog} which we discuss in the next section.

Tagalog has a non-configurational phrase structure \citep{kroeger1993phrase} wherein sentences are canonically predicate-initial and sentence arguments following the predicate have flexible order and can also be fronted using the \textit{ay} inversion marker.
Tagalog lacks a copula, and thereby permits noun phrases, prepositional phrases, and adjectival forms to be sentence predicates \citep{reid2005tagalog}; specificational and identificational clauses in Tagalog are formed by a simple juxtaposition of the subject and complement.
The language is also noted for its productive affixation, which allows lexical terms from typologically dissimilar languages like English to be easily encoded in its morphosyntax \citep{tangco2002taglish}, and its varied forms and functions of reduplication \citep{blake1917reduplication}.

\subsection{UD annotation of Tagalog}
\label{sec:challenges}
Despite extensive linguistic scholarship on Tagalog, some features of Tagalog grammar have been the topic of continued debate among linguists \cite{javier2022tagalog}.
In annotating the \udnewscrawl{} treebank, we made decisions which reveal our positions in some of these debates.
Here we discuss our approaches to two major points of contention in the syntactic analysis of Tagalog; several other annotation choices dealing with specifics of the UD framework are outlined in Appendix \ref{appendix:annotation_guidelines}.
The possible typological implications of these choices are outside the scope of this paper but receive a more thorough treatment in \citet{bardaji2024}.

\paragraph{Voice marking system.}
One debate in Tagalog grammar is related to its voice marking system and assignment of grammatical relations \citep{Cubar1975,Schachter-1976,Rafael2016}.
Tagalog and other ``Philippine-type languages'' are said to have a distinctive voice marking system, wherein essentially any type of clausal argument\textemdash which may be a noun phrase (NP) referring to an entity or object that performs the thematic role of agent, patient, location, beneficiary, or instrument\textemdash can be marked as subject depending on the voice affix attached to the predicate head.
In \autoref{fig:examples}, the NP preceded by the marker \textit{ang} is generally viewed as the subject in the sentence and agrees with the voice marking on the predicate head, while the markers \textit{ng} and \textit{sa} precede non-subject arguments and adjuncts. These markers have respective counterparts for marking proper nouns.

For this treebank annotation project, we adopted a \textbf{symmetrical voice} view of Tagalog voice marking.
Following this view, we consider all three sentences in \autoref{fig:examples} as ``unmarked'' or basic transitive sentences, as such languages do not show a preference for the agent argument to be the subject, unlike in Indo-European languages like English \citep{Foley2008,Riesberg2019}.

In \autoref{fig:examples} therefore, while the verbs in each sentence exhibit different voice markings and the subject carry different thematic roles, the examples could all be translated as the equivalent of the English sentence ``The man gave flowers to the woman,'' given that no process of agent demotion takes place in Tagalog constructions where the agent is not the focused argument, which is contrary to what would happen in English passive voice constructions.
However, it should also be noted that a speaker's choice of voice is usually affected by different factors, which could include definiteness \citep{Himmelmann2005}, specificity \citep{Rackowski2005}, and topicality in discourse \citep{duncan1985}.

The type of voice marking used in a clause is indicated at the morphological features level of our UD annotation scheme, which is similar to how they are marked in the UD-TRG treebank.
The following voice types were marked in \udnewscrawl{}:
(1) \texttt{Act}, which is the label UD assigns to both the Indo-European active voice and the actor-focus voice of Austronesian languages;
(2) \texttt{Pass}, which is short for passive but which also applies to the Austronesian patient-focus voice,
(3) \texttt{Bfoc} for beneficiary-focus voice,
(4) \texttt{Lfoc} for location-focus voice, and
(5) \texttt{Cau} for causative forms which UD classifies as a voice category.

In indicating the dependency relation between the predicate head and its core arguments, we met theoretical issues regarding the working definitions of certain categories in the UD framework.
For example, the \texttt{nsubj} relation is defined in terms of the semantic role carried by the subject NP and favors grammatical systems such as those seen in Indo-European languages like English where the agent is usually seen as the more privileged argument.
While such UD labels can be edited to a certain extent in the language-specific documentation to better suit the grammar of a language, incompatibilities such as in the example provided reveals how English has become the \textit{de facto} standard for understanding how languages work.

{
\setlength{\tabcolsep}{2pt}
\begin{table*}[t]
    \centering
    \resizebox{\textwidth}{!}{%
        \begin{tabular}{@{}lcccrr@{}}
            \toprule
                                                           & \multicolumn{3}{c}{\textbf{\# Sents / Tokens}} & \multicolumn{2}{c}{\textbf{\# Tags}}                                  \\
            \textbf{Treebank}                              & Train                                          & Dev                                  & Test           & UPOS & DEPREL \\ \midrule
            \textbf{\udnewscrawl{}}                        & 12.4k / 286.9k                                 & 1.56k / 37.0k                        & 1.56k / 36.9k  & 17   & 39     \\
            UD-TRG \citep{Samson2018TRG}                   & $-$                                            & $-$                                  & 128 / 734      & 13   & 26     \\
            UD-Ugnayan \citep{aquino-de-leon-2020-parsing} & $-$                                            & $-$                                  & ~~~~94 / 1.01k & 14   & 24     \\ \bottomrule
        \end{tabular}
    }
    \caption{Dataset statistics for \udnewscrawl{} and its comparison to existing Tagalog treebanks in the Universal Dependencies (UD) framework.
        The breakdown of tags per split is found in Appendix \ref{appendix:breakdown}. 
    }
    \label{tab:dataset_details}
\end{table*}
}

\paragraph{Categorization of Tagalog roots.}
The lexical categories of Tagalog roots and the question of whether or not they should be considered pre-categorial prior to affixation have also been a topic of debate among linguists \citep[see][]{De-Guzman-1978,Himmelmann1991,Himmelmann2005,Kaufman2009}.
\citet{Himmelmann2007}, for example, observed that in several Tagalog dictionaries, many roots that could be presumed verbal are glossed with English nouns or adjectives, such as `gift' for \textit{bigay} and `surpassed, defeated' for \textit{daig}, even though these can be inflected with verb affixes, thus they can alternatively be glossed as 'to give' and 'to surpass or defeat', respectively.

The POS designation of Tagalog roots in \udnewscrawl{} was determined by the morphological structure of the word.
Thus, if a root is affixed with a nominal affix, such as \textit{pag-} in the word \textit{paglakad} (manner of walking), then it is labeled as \texttt{NOUN}.
However, if a verbal affix is attached to it, such as \textit{mag-} in \textit{maglakad}  'to walk', then it is labeled as \texttt{VERB}.
Meanwhile, the POS of words that are unaffixed or appear in their bare forms are usually determined by their syntactic distribution.
Otherwise, the Tagalog-English dictionary of Leo James English was consulted if the POS of a word remains ambiguous to the manual annotator.

\section{The \udnewscrawl{} treebank}
\label{sec:udnewscrawl}

\udnewscrawl{} contains 15.6k sentences (360.8k tokens) manually annotated by Tagalog native speakers.
Individual words in the dataset underwent lemmatization and POS tagging.
Dependency relations of words in each sentence were identified, and select morphological features were also annotated.
Dataset details can be found in \autoref{tab:dataset_details}.
In this section, we describe the context of the annotation project, the nature of text included in the corpus, and our procedures for annotation and quality control.

\subsection{Project context}
\udnewscrawl{} was originally developed to fulfill a different linguistic analysis objective, i.e., to investigate cross-linguistic variation in the distribution of lexical information, especially on languages like Tagalog that exhibit different degrees of formal distinction between their syntactic categories (\S\ref{sec:challenges}).
As a result, there were annotation decisions incompatible with the original UD guidelines (as seen in Appendix \ref{appendix:annotation_guidelines}).
After creating the initial version of the treebank, we perform post-processing to ensure that it meets current UD standards (\S\ref{sec:qualitycontrol}).

We recruited a total of 15 annotators to manually label \udnewscrawl{} using the WebAnno 3.6.4 annotation platform.
All annotators are native speakers of Tagalog, with most annotators having an undergraduate to postgraduate level of linguistics education.
Annotators were given prior training on linguistic concepts involved in UD syntactic analysis, as well as the use of the WebAnno platform.
They were then compensated at a rate of \PHP2500 (roughly €40) for each set of 100 sentences annotated, with an estimated workload of 20 hours per set.
The annotation project spanned 16 months, including planning and actual annotation, while the quality control process lasted for a year.

\begin{figure*}
  \centering
  \includegraphics[width=\linewidth]{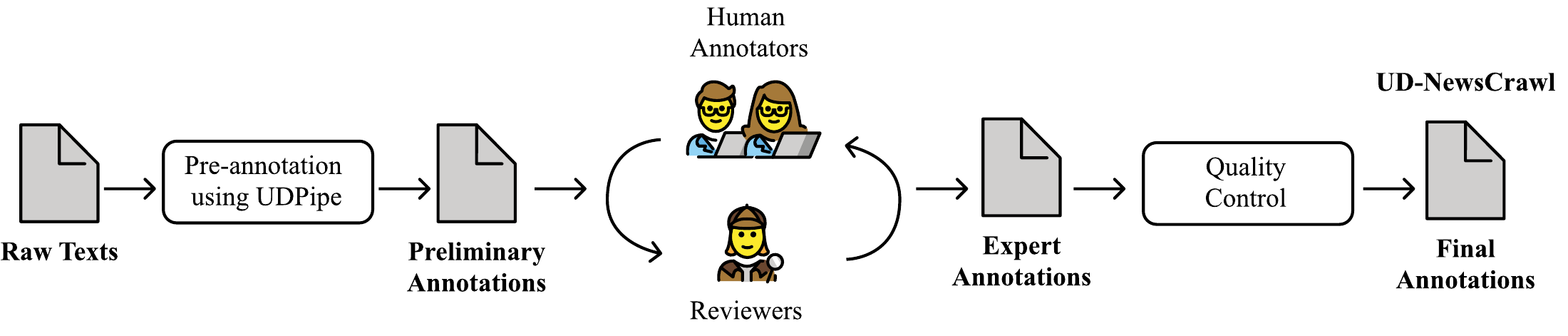}
  \caption{Annotation workflow for \udnewscrawl{}.}
  \label{fig:iterative_annotation}
\end{figure*}

\subsection{Text corpus properties}
The text included in the treebank is sourced from the Leipzig Tagalog NewsCrawl 2011 corpus, which consists of material collected from Tagalog news sites \citep{quasthoff1998projekt,goldhahn-etal-2012-building}.
Sample sentences from the corpus can be found in Appendix \ref{appendix:topic_examples}, with most sentences written in the declarative and more formal register of Tagalog.
Some use of informal language can also be found in the corpus, particularly when texts quote a person's speech.
We also find several instances of code-switching in the texts, reflecting the bilingual nature of many Tagalog speakers.

The data collection team automated the sentence tokenization of texts in the corpus and selected a subset of 15,000 sentences.
These were then divided into 150 files of 100 sentences each in order of increasing sentence length, with an approximately even distribution of sentence lengths across files.
This data preparation process distributed the annotation difficulty across files, and allowed an annotator working on one file to start with shorter, simpler sentences then progress to longer, more challenging sentences as their proficiency in the annotation workflow increased.
However, the data preparation process also resulted in sentences being isolated from their original context, and some errors in the automated sentence tokenization produced fragments or run-on sentences instead, which made the subsequent annotation more challenging.

\subsection{Annotation procedure}
An initial set of annotation guidelines was prepared based on the UD framework and the project's analysis objectives.
We then employed an iterative annotation workflow (c.f. \citet{Fort2016CollaborativeAF}) for \udnewscrawl{}, as seen in \autoref{fig:iterative_annotation}, which enabled us to modify and expand on the annotation guidelines as we encountered cases in the texts which were not previously covered.
We highlight aspects of our annotation guidelines in Appendix \ref{appendix:annotation_guidelines}.

\paragraph{Pre-annotation.}
Prior to human annotation, POS tags and lemmas were annotated using a UDPipe model trained on the UD-Ugnayan treebank, resulting in a set of \textbf{preliminary annotations}.
This reduced the cognitive load of annotators who only needed to validate the pre-annotated labels and correct as needed instead of labeling from scratch.

We opted not to pre-annotate morphological features and dependency relations, since the higher complexity of word forms and sentence structures found in the NewsCrawl corpus resulted in a high error rate for these annotations from models trained on existing Tagalog treebanks with simpler texts.

\paragraph{Human annotation.}
Annotators were tasked to first validate and correct tokenization, especially the splitting of linkers and contractions.
Once the tokenization was validated, annotators proceeded to label POS tags, dependency relations, and morphological features following the working annotation guidelines.
Because of differences in annotators' competency in linguistic annotation, the annotation workload for some files was divided between two annotators: less experienced annotators handled tokenization and POS tagging while more experienced annotators were tasked with labeling dependency relations and morphological features.
Finally, we encouraged annotators to keep a set of notes to document edge cases and unusual examples for further deliberation.

\paragraph{Verification and re-annotation.}
A senior linguist independently reviewed the annotators' work to validate decisions and identify potential inconsistencies.
We discussed and resolved difficult annotation cases in regular meetings, and updated the annotation guidelines accordingly.
Cases requiring revision were re-annotated following the updated annotation guidelines.
This step helped maintain quality and consistency across the dataset.
This cycle of annotation, verification, and re-annotation resulted in a set of \textbf{expert annotations} that we send for quality control.

{
\begin{table*}[t]
    \centering
    \resizebox{\textwidth}{!}{%
        \begin{tabular}{@{}lrrrrrr@{}}
            \toprule
            \textbf{Pipeline}                                         & \textbf{Lemm.}                & \textbf{UPOS}                 & \multicolumn{2}{c}{\textbf{Morph.}} & \multicolumn{2}{c}{\textbf{Dep. Parsing}}                                                                 \\
            \textit{Feature representation}                           & \textit{Acc}                  & \textit{Acc}                  & \textit{Acc}                        & \textit{F1-score}                         & \textit{UAS}                  & \textit{LAS}                  \\ \midrule
            No embeddings                                             & 89.5{\small$\pm$1.1}          & 89.7{\small$\pm$0.8}          & 94.3{\small$\pm$0.8}                & 92.8{\small$\pm$0.4}                      & 82.4{\small$\pm$1.0}          & 75.4{\small$\pm$0.8}          \\
            \hdashline
            \textit{Word embeddings}                                                                                                                                                                                                                                                    \\
            fastText \citep{bojanowski-etal-2017-enriching}           & 89.8{\small$\pm$0.8}          & 90.3{\small$\pm$0.3}          & 94.9{\small$\pm$0.3}                & 93.9{\small$\pm$0.2}                      & 83.1{\small$\pm$0.5}          & 76.2{\small$\pm$0.5}          \\
            Multi hash embeddings \citep{miranda2022multi}            & 90.3{\small$\pm$0.8}          & 90.9{\small$\pm$0.1}          & 95.4{\small$\pm$0.1}                & 94.6{\small$\pm$0.1}                      & 83.9{\small$\pm$0.5}          & 77.4{\small$\pm$0.6}          \\
            \hdashline
            \textit{Context-sensitive vectors}                                                                                                                                                                                                                                          \\
            mDeBERTa-v3, base \citep{he2021deberta}                   & 90.6{\small$\pm$1.3}          & 91.3{\small$\pm$0.8}          & 95.2{\small$\pm$0.7}                & 94.7{\small$\pm$0.7}                      & 85.1{\small$\pm$0.8}          & 79.0{\small$\pm$0.6}          \\
            RoBERTa-Tagalog, large \citep{cruz-cheng-2022-improving}  & 90.4{\small$\pm$0.9}          & \textbf{91.7{\small$\pm$0.7}} & \textbf{95.6{\small$\pm$0.8}}       & \textbf{95.1{\small$\pm$0.6}}             & 86.4{\small$\pm$0.7}          & 80.6{\small$\pm$0.6}          \\
            XLM-RoBERTa, large \citep{conneau-etal-2020-unsupervised} & \textbf{91.0{\small$\pm$0.9}} & 91.5{\small$\pm$0.9}          & 95.4{\small$\pm$0.8}                & \textbf{95.1{\small$\pm$0.6}}             & \textbf{86.9{\small$\pm$0.0}} & \textbf{81.0{\small$\pm$0.1}} \\ \bottomrule
        \end{tabular}
    }
    \caption{
        Test set performance on various linguistic tasks using the \udnewscrawl{} given different feature representations.
        We report the average of three runs and their standard deviation.
        Full results can be found in Appendix \ref{appendix:full_results}.
    }
    \label{tab:baseline_results}
\end{table*}
}

\subsection{Treebank quality-control}
\label{sec:qualitycontrol}

To ensure that the expert annotations remained consistent throughout the treebank, we employed both semi-automated and manual post-annotation workflows.
First, we trained a silver-standard parsing model on our existing annotations using spaCy \citep{spacy} and identified instances where the model's morphological annotations and dependency parsing relations disagree with human annotations.
This approach is based on the premise that a silver-standard model can learn global patterns from the training labels, even when trained on partially incorrect data \citep{tedeschi-etal-2021-wikineural-combined,zhang-etal-2022-survey,wang-etal-2024-use}.
These cases were prioritized for review, as disagreements often indicated potential inconsistencies or errors.
Second, we conducted manual quality checks through random sampling of sentences, comparing them against established UD treebanks using the UD official validator\footnote{\url{https://github.com/UniversalDependencies/tools}} to ensure adherence to UD guidelines and consistency.
We describe the results of this error analysis in Appendix \ref{appendix:error_analysis}.
After quality control, we obtain a set of \textbf{final annotations} which forms the basis for \udnewscrawl{} and succeeding experiments.

\section{Baseline Models for \udnewscrawl{}}
\label{sec:baseline}

\paragraph{Set-up.}
To establish baseline performance for \udnewscrawl{}, we trained several multi-task models that accommodate different compute requirements using the spaCy framework \citep{spacy}.
Each model consists of the following components (full description in Appendix \ref{appendix:components}):

\begin{itemize}
  \setlength\itemsep{-0.2em}
  \item \textbf{Lemmatizer}: employs an edit-tree recursive algorithm based on \citet{muller-etal-2015-joint}. We train a convolutional network with a softmax layer to predict the best edit-tree for a token.
  \item \textbf{Morphological and UPOS tagger}: treats morphological annotation and POS tagging as a multilabel classification problem and implements a softmax layer to predict scores given a token.
  \item \textbf{Dependency parser}: uses a variant of the non-monotonic arc-eager transition system as described in \citet{honnibal-johnson-2015-improved}.
\end{itemize}

In order to investigate how different feature representations affect model performance, we trained these components using several feature representations:
(1) fastText word embeddings \citep{bojanowski-etal-2017-enriching}, (2) spaCy's multi hash embeddings as described in \citet{miranda2022multi}, (3) monolingual context-sensitive vectors using RoBERTa Tagalog \citep{cruz-cheng-2022-improving}, and (4) multilingual context-sensitive vectors using XLM-RoBERTa \citep{conneau-etal-2020-unsupervised} and mDeBERTa-v3 \citep{he2021deberta}.
For all context-sensitive vectors, we use the \texttt{large} variant of the pretrained models except for mDeBERTa-v3, which is not available.
The full training hyperparameters can be found in Appendix \ref{appendix:hyperparameters}.
Finally, we evaluated each component in its corresponding linguistic task.
We report the accuracy for both lemmatization and POS tagging tasks and the macro F1-score for the morphological annotation task.
In addition, we also report the unlabeled and labeled attachment scores (UAS / LAS) for the dependency parsing task.

\paragraph{Results.}
\autoref{tab:baseline_results} presents the performance of various baseline models trained on \udnewscrawl{}.
The model trained on multilingual XLM-RoBERTa context-sensitive vectors achieves the best performance in most tasks, with 1--2\% (absolute) improvement from the ``No embeddings'' baseline.
This performance is closely followed by monolingual RoBERTa, which was trained specifically on Tagalog texts.
We hypothesize that XLM-RoBERTa leverages cross-lingual transfer \citep{artetxe2019cross}, allowing it to perform at par with a fully-monolingual model.

\section{Analysis}
We perform several analyses to evaluate the quality of expert annotations (\S\ref{sec:quality}), the generalization of \udnewscrawl{} to other Tagalog treebanks (\S\ref{sec:crosstreebank}), and the type of content that exists within the treebank (\S\ref{sec:topic_clf}).

\subsection{Quality analysis}
\label{sec:quality}

\begin{table}[t]
    \centering
    \begin{tabular}{lrr}
        \toprule
        \textbf{Metric} & \textbf{Cohen's $\kappa$} & \textbf{\% Corrected} \\
        \midrule
        UPOS            & 0.75                      & 15\%                  \\
        Dep. Rel.       & 0.68                      & 20\%                  \\
        Morph.          & 0.70                      & 22\%                  \\
        \bottomrule
    \end{tabular}
    \caption{Initial disagreement between the silver-standard model and original annotations, and the proportion of disagreed sentences corrected.}
    \label{table:disagreement}
\end{table}

\paragraph{Set-up.}
During the quality control process (\S\ref{sec:qualitycontrol}), we developed a silver-standard model to compare its annotations against expert annotations.
We focused on identifying and correcting sentences where discrepancies occurred.
To evaluate the level of agreement, we used Cohen's $\kappa$ and calculated the proportion of sentences corrected out of all the cases with disagreements.

\paragraph{Results.}
\autoref{table:disagreement} shows the initial disagreement, as measured by Cohen's $\kappa$ between the silver-standard model and the original annotations, together with the proportion of the disagreed sentences that we corrected.
The results indicate moderate agreement between the silver-standard model and the original annotations \citep{mchugh2012interrater}.
These results imply that while the annotations are fairly consistent, there is room for improvement, especially in dependency relations and morphological features.
The percentage of corrected sentences shows that a small portion of the annotations required adjustments, highlighting areas where the silver-standard model's predictions diverged from human annotations.

\subsection{Cross-treebank generalization on UD-TRG and UD-Ugnayan}
\label{sec:crosstreebank}

{
\begin{table*}[t]
    \centering
    \begin{tabular}{@{}lrrrrrr@{}}
        \toprule
                                              & \textbf{Lemm.}                & \textbf{UPOS}                 & \multicolumn{2}{c}{\textbf{Morph.}} & \multicolumn{2}{c}{\textbf{Dep. Parsing}}                                                                 \\
        \textbf{Treebank}                     & \textit{Acc}                  & \textit{Acc}                  & \textit{Acc}                        & \textit{F1-score}                         & \textit{UAS}                  & \textit{LAS}                  \\ \midrule
        \textit{UD-TRG}                                                                                                                                                                                                                                         \\
        \textbf{Ours}                         & \textbf{81.1{\small$\pm$0.4}} & \textbf{78.3{\small$\pm$0.6}} & \textbf{73.4{\small$\pm$0.4}}       & \textbf{78.5{\small$\pm$0.4}}             & \textbf{95.5{\small$\pm$0.2}} & \textbf{68.5{\small$\pm$0.4}} \\
        \citet{dehouck-denis-2019-phylogenic} & $-$                           & $-$                           & $-$                                 & $-$                                       & 70.9                          & 50.4                          \\
        \citet{kondratyuk-straka-2019-75}     & 75.0                          & 61.6                          & 35.3                                & $-$                                       & 64.7                          & 39.4                          \\
        \hdashline
        \textit{UD-Ugnayan}                                                                                                                                                                                                                                     \\
        \textbf{Ours}                         & 83.3{\small$\pm$0.4}          & \textbf{82.3{\small$\pm$0.4}} & $-$                                 & $-$                                       & \textbf{82.8{\small$\pm$0.4}} & \textbf{60.8{\small$\pm$0.4}} \\
        \citet{aquino-de-leon-2020-parsing}   & \textbf{85.5}                 & 80.5                          & $-$                                 & $-$                                       & 63.5                          & 55.4                          \\ \bottomrule
    \end{tabular}
    \caption{
        Performance comparison of our best performing pipeline (\textbf{Ours}), i.e., context-sensitive vectors from XLM-RoBERTa trained on \udnewscrawl{}, on the UD-TRG and UD-Ugnayan treebanks against previous reported results in literature.
        Reporting the average of three runs and the standard deviation for our results.
    }
    \label{tab:cross_treebank_results}
\end{table*}
}

\paragraph{Set-up.}
In order to understand how well the parsers trained on \udnewscrawl{} generalize to other datasets, we used the best performing model\textemdash i.e., with components trained on context-sensitive vectors from XLM-RoBERTa\textemdash and evaluate it on the UD-TRG and UD-Ugnayan treebanks.
We also compared against the best reported results from the literature for both treebanks such as \citet{dehouck-denis-2019-phylogenic}'s phylogenic tree approach and \citet{kondratyuk-straka-2019-75}'s UDify for UD-TRG, and \citet{aquino-de-leon-2020-parsing}'s cross-lingual approach for UD-Ugnayan.

\paragraph{Results.}
Our results in  \autoref{tab:cross_treebank_results} suggest that UD-NewsCrawl effectively captures the linguistic structures and patterns of Tagalog, making it a robust resource for dependency parsing.
However, we note that the results are severely affected by the small size of the other treebanks, which hinders a proper comparison.
Despite this, the models achieve state-of-the-art performance by surpassing previously reported benchmarks on both treebanks.
The cross-treebank generalization indicates that despite potential differences in domain, annotation style, or text genre, 
the syntactic patterns encoded in these models are potentially transferrable across various contexts.

\subsection{Topics in \udnewscrawl{}}
\label{sec:topic_clf}

\paragraph{Set-up.}
In order to identify the topics present in \udnewscrawl{}, we performed zero-shot classification using Llama-3.1-8B-Instruct \citep{dubey2024llama} given a set of topics defined in SIB-200 \citep{adelani-etal-2024-sib} (prompt template can be found in Appendix \ref{appendix:prompt_topic}), then manually evaluated the results.
\autoref{fig:topic_clf} shows the topic distribution for \udnewscrawl{}.
Examples for each topic can be found in Appendix \ref{appendix:topic_examples}.
In addition, we also perform a more fine-grained topic analysis as seen in Appendix \ref{appendix:fine_grained_topic_clf}

\paragraph{Results.}
Our analysis of the topic distribution reveals a clear dominance of entertainment and health-related content, constituting over 60\% of the dataset.
This skewed distribution highlights potential collection biases in web-crawled news data, which may impact downstream applications relying on this dataset for training or evaluation.

\begin{figure}[t]
  \centering
  \includegraphics[height=0.95\linewidth]{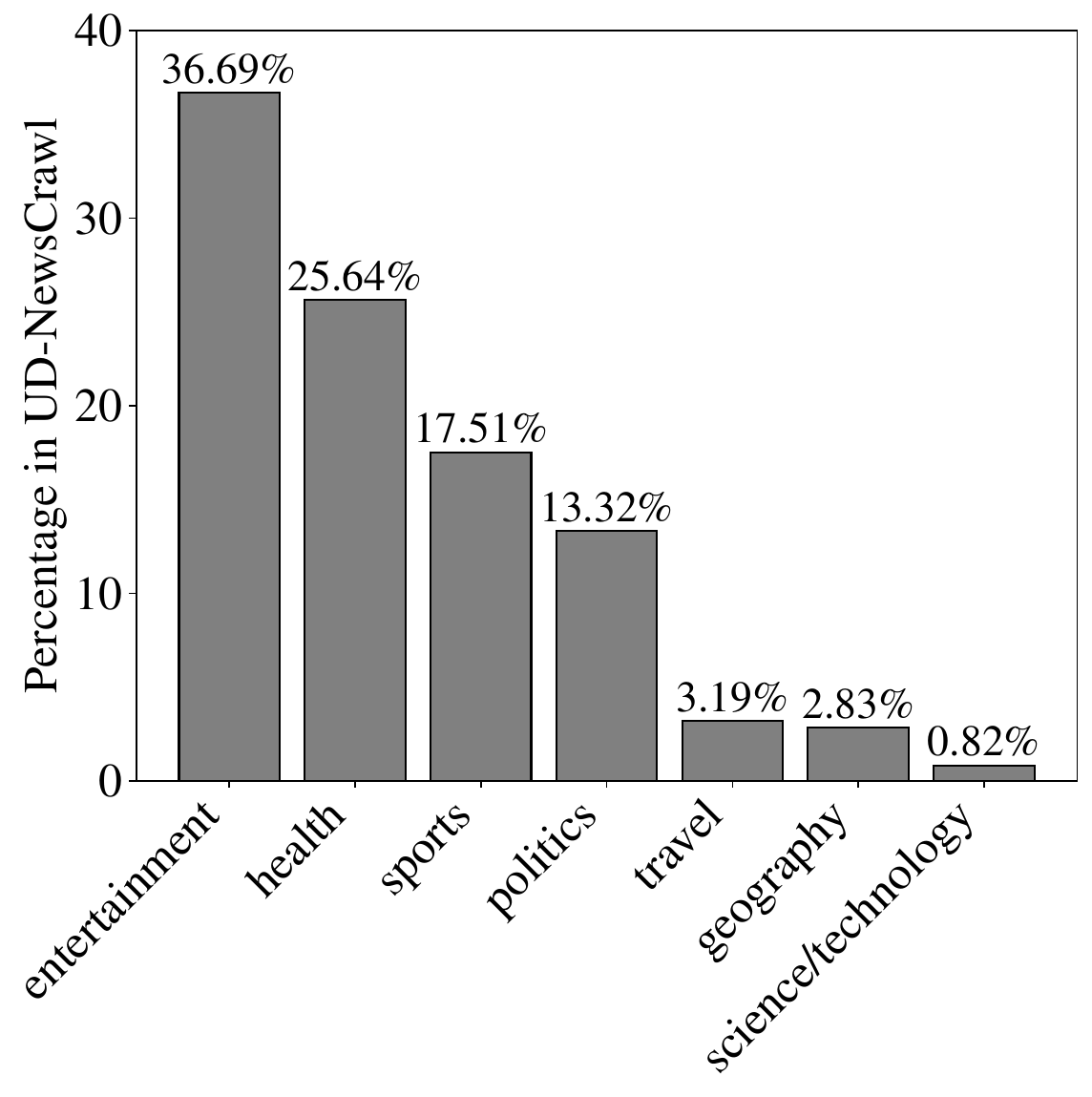}
  \caption{Topic distribution of \udnewscrawl{} using categories from SIB-200 \citep{adelani-etal-2024-sib}.}
  \label{fig:topic_clf}
\end{figure}

\section{Discussion}

\paragraph{Syntactic annotation in a time of LLMs.}
One key motivation for training dependency parsers (and consequently, creating treebanks) has been their downstream utility in NLP applications such as question answering, grammar rule extraction, and semantic role labeling \citep[\textit{inter alia}]{liang-etal-2011-learning, berant-etal-2013-semantic,herrera-etal-2024-sparse}.
However, the emergence of large language models (LLMs) that can directly perform many of these tasks brings into question the continued relevance of labor-intensive treebank creation.

We argue that syntactic annotations remain valuable not just as training data, but as \textbf{explicit, interpretable representations of linguistic structure} that enable detailed analysis of language phenomena, hypothesis testing about grammar, and evaluation of model capabilities in ways that end-task performance alone cannot capture.
As noted by \citet{lappin2024assessing}, LLMs are biased towards certain language expressions by both their architecture and represented distributions in the training data \citep[\textit{inter alia}]{ryan-etal-2024-unintended, bhatt-diaz-2024-extrinsic}.
Additionally, LLMs are opaque as we cannot fully explain their representations of language nor reliably modify them to produce set outcomes.


\paragraph{Challenging the universality of UD.}
While Universal Dependencies aims to provide a cross-linguistically consistent annotation scheme, several linguistic phenomena challenge its universality \citep{osborne2019status, kanayama-iwamoto-2020-universal}.
For example, in \udnewscrawl{}, the \texttt{nsubj} relation is defined in terms of subject NP semantic roles typical of Indo-European languages.
Although UD provides language-specific documentation, English (and Euro-western linguistic frameworks) remains the default reference point for analyzing syntactic structures.

We highlight the importance for NLP practitioners to reflect on the implicit biases inherent in current linguistic frameworks and model architectures, which are often initialized and optimized based on globally dominant languages.
This reflection is crucial for developing more equitable and effective NLP tools that respect and accommodate linguistic diversity.

\section{Conclusion}
In this work, we introduced \udnewscrawl{}, the largest Tagalog treebank to date, consisting of 15,619 sentences manually annotated according to the Universal Dependencies framework.
We described our development process for \udnewscrawl{}, which involved multiple stages of manual and semi-automated annotation and verification to ensure the quality of the treebank.
We also discussed several linguistic challenges specific to Tagalog which required careful consideration during the annotation process.
Our baseline evaluations and experiments using transformer-based models trained on \udnewscrawl{} demonstrate the robustness and generalizability of the dataset, highlighting its utility in different areas of computational linguistics research and development.

We anticipate that \udnewscrawl{} will serve as a valuable resource for researchers and practitioners working on Tagalog NLP, enabling the development of more accurate and robust language technologies.
We hope that insights gained from this project can also inform future efforts to create syntactic resources for other low-resource languages, contributing to a more inclusive and diverse NLP landscape.

\section*{Author contributions}
Or led the annotation project, with Aquino as the technical consultant for UD annotation.
Miranda assisted with the treebank quality-control and on training baselines from the treebank.
All authors contributed in drafting and writing the manuscript.

\section*{Limitations}

\paragraph{Domain and topic bias}
The corpus is primarily composed of news articles, which may introduce domain-specific biases.
This could limit the generalizability of models trained on \udnewscrawl{} to other domains, such as conversational or literary texts.
In addition, the corpus is skewed towards entertainment and health-related content, which may not fully represent the diversity of topics in Tagalog.
This could impact the performance of models in downstream applications that require broader topic coverage.
Finally, the timeframe of the texts extends until 2011, potentially missing more recent developments in language use and contemporary topics.
This temporal limitation may affect the model's ability to process modern terminology and current cultural references, particularly in rapidly evolving domains such as technology and social media.

\paragraph{Annotation challenges}
Despite rigorous quality control, the annotation process faced challenges due to the context of the project.
For example, a portion of the annotation work were done during the height of the COVID-19 pandemic (from January 2021 to April 2022), making it difficult for some annotators to stay throughout the project.
Some inconsistencies may still exist despite quality-control, especially in edge cases.

\section*{Ethics Statement}
The development of \udnewscrawl{} involved the manual annotation of texts by native Tagalog speakers, who were compensated for their work.
We ensured that the annotators were treated fairly and that their contributions were acknowledged.
The data used in the corpus was sourced from publicly available news articles, and we adhered to ethical guidelines regarding data usage and privacy.
However, it is important to consider the potential ethical implications of using web-crawled data, particularly in terms of copyright and the representation of diverse voices.
While the corpus reflects the bilingual nature of many Tagalog speakers, it may not fully capture the linguistic diversity of the Philippines, which is home to over 170 languages.
Future efforts should aim to include a more diverse range of texts and dialects to ensure broader representation.
Finally, the creation of \udnewscrawl{} is intended to support the development of NLP applications for Tagalog, a language that has been historically underrepresented in computational linguistics.
By providing this resource, we hope to contribute to the preservation and promotion of Tagalog in the digital age, while also encouraging similar efforts for other low-resource languages.

\section*{Acknowledgements}
The treebank was originally created for the Deutsche Forschungsgemeinschaft (DFG)-funded project entitled ``Information distribution and language structure - correlation of grammatical expressions of the noun/verb distinction and lexical information content in Tagalog, Indonesian and German'' which was headed by Gerhard Heyer from Leipzig University and Nikolaus P. Himmelmann from the University of Cologne.
The authors would also like to thank all annotators from the Linguistics Department of the University of the Philippines - Diliman, who helped in the project: Patricia Anne Asuncion, Paola Ellaine Luzon, Jenard Tricano, Mary Dianne Jamindang, Michael Wilson Rosero, Jim Bagano, Yeddah Joy Piedad, Farah Cunanan, Calen Manzano, Aien Gengania, Prince Heinreich Omang, Noah Cruz, Leila Ysabelle Suarez, Orlyn Joyce Esquivel, and Andre Magpantay.
Finally, we would also like to thank the reviewers during the February ARR cycle for their thoughtful comments and feedback that helped improve this paper.

\bibliography{custom}

\clearpage
\appendix

\section{Tag breakdown for \udnewscrawl{}}
\label{appendix:breakdown}

\autoref{tab:upos_tags} and \autoref{tab:deprel_tags} shows the distribution of Universal Part-of-Speech (UPOS) and Dependency Relations (DEPREL) tags for all splits of \udnewscrawl{}.

{
\setlength{\tabcolsep}{2pt}
\begin{table}[h]
    \centering
    \begin{tabular}{lrrr}
        \toprule
              & \textbf{Train} & \textbf{Dev} & \textbf{Test} \\
        \midrule
        NOUN  & 41722          & 5450         & 5543          \\
        ADP   & 33003          & 4207         & 4292          \\
        PROPN & 32034          & 3929         & 3858          \\
        VERB  & 28361          & 3802         & 3668          \\
        PART  & 27899          & 3673         & 3790          \\
        PUNCT & 23384          & 2950         & 2978          \\
        ADV   & 21029          & 2797         & 2718          \\
        DET   & 19515          & 2582         & 2503          \\
        PRON  & 15555          & 1954         & 1899          \\
        ADJ   & 9173           & 1099         & 1132          \\
        CCONJ & 6621           & 891          & 852           \\
        SCONJ & 6254           & 794          & 762           \\
        NUM   & 6088           & 756          & 790           \\
        INTJ  & 175            & 34           & 31            \\
        SYM   & 122            & 32           & 19            \\
        X     & 56             & 9            & 7             \\
        \bottomrule
    \end{tabular}
    \caption{
        UPOS (Universal Part-of-Speech) distribution
    }
    \label{tab:upos_tags}
\end{table}
}
{
\setlength{\tabcolsep}{2pt}
\begin{table}[ht!]
    \centering
    \begin{tabular}{lrrr}
        \toprule
                       & \textbf{Train} & \textbf{Dev} & \textbf{Test} \\
        \midrule
        case           & 45243          & 5800         & 5870          \\
        punct          & 23360          & 2948         & 2973          \\
        det            & 19801          & 2632         & 2568          \\
        advmod         & 19690          & 2610         & 2529          \\
        nsubj          & 18322          & 2389         & 2281          \\
        nmod           & 16012          & 1997         & 2074          \\
        mark           & 15375          & 2031         & 2032          \\
        flat           & 14925          & 1814         & 1782          \\
        obl            & 12953          & 1691         & 1735          \\
        root           & 12495          & 1561         & 1563          \\
        nmod:poss      & 7780           & 1025         & 1039          \\
        advcl          & 7113           & 917          & 840           \\
        conj           & 6865           & 947          & 886           \\
        cc             & 6497           & 860          & 829           \\
        obj:agent      & 6239           & 765          & 735           \\
        amod           & 6131           & 706          & 773           \\
        acl:relcl      & 5359           & 715          & 743           \\
        obj            & 4879           & 583          & 600           \\
        discourse      & 3622           & 492          & 501           \\
        fixed          & 3492           & 487          & 503           \\
        nummod         & 3191           & 417          & 401           \\
        ccomp          & 2982           & 362          & 392           \\
        xcomp          & 1722           & 293          & 241           \\
        compound       & 1464           & 286          & 307           \\
        parataxis      & 1340           & 165          & 164           \\
        appos          & 1334           & 176          & 187           \\
        dep            & 722            & 71           & 81            \\
        dislocated     & 710            & 84           & 79            \\
        compound:redup & 439            & 44           & 49            \\
        list           & 322            & 17           & 16            \\
        acl            & 317            & 51           & 50            \\
        vocative       & 113            & 13           & 15            \\
        goeswith       & 79             & 6            & 3             \\
        orphan         & 77             & 2            & 1             \\
        reparandum     & 10             & 1            & 0             \\
        iobj           & 9              & 0            & 0             \\
        obl:agent      & 6              & 0            & 0             \\
        cop            & 1              & 0            & 0             \\
        csubj          & 1              & 0            & 0             \\
        expl           & 1              & 0            & 0             \\
        \bottomrule
    \end{tabular}

    \caption{
        DEPREL (Dependency relations) distribution
    }
    \label{tab:deprel_tags}
\end{table}
}

\section{Annotation Process}
\label{appendix:annotation_guidelines}

\subsection{Annotation Guidelines}
We highlight some aspects of the annotation guidelines in this section.
The full annotation guidelines will be made available after the review period.

\subsubsection{Tokenization and lemmatization}

\paragraph{Multiword tokens.}
Since the basic units of annotation in UD are syntactic words, we systematically split off clitics in case they are written attached to their host.
In Tagalog, this only concerns \textbf{multiword tokens}, in which the linker \textit{-ng} when it is attached to a vowel-final word.

\paragraph{Foreign words.}
For English or other foreign words mixed in the data, we do not identify the root words.
Exceptions are when they are used with Tagalog affixes:

\begin{itemize}
  \setlength\itemsep{-0.5em}
  \item \textit{nag-e-enjoy} (enjoying) \textrightarrow~enjoy
  \item \textit{nakipag-meeting} (had a meeting) \textrightarrow~meeting
  \item \textit{kaka-check} (just checked, checking) \textrightarrow~check
\end{itemize}

\paragraph{On some multiword expressions (MWEs).}
In Tagalog, some MWEs are now commonly contracted into a single word, such as \textit{kundi} (previously \textit{kung hindi}, ``if not'') or \textit{anuman} (previously \textit{ano man}, ``whatever'').
We have opted not to separate these expressions further, and we instead retain them as individual words with lemmas as they were written in the text.

\subsubsection{POS Tags}

In Tagalog, POS tagging can be tricky and confusing.
In general, the POS of words are assigned based on the derived word rather than the root.
For example, in \textit{paglakad} (Lemma=\textit{lakad}, ``walk''), the root \textit{lakad} may be used as a noun or as a verb.
But, if combined with the affix \textit{pag-},  then it can only be used as a noun.
For words that appear in their bare forms, we consult the Tagalog-English Dictionary \citep{english1986tagalog}.
In general, if a word denotes an object, then they are labelled as a \texttt{NOUN}, \texttt{ADJ}, or \texttt{VERB}.

\subsubsection{Morphological Features}

In \udnewscrawl{}, we mark the morphological features that affixes add to the word.
Hence, in the sentence \textit{``Lakad na tayo.''} (Let's walk), there is no need to mark any features on \textit{lakad} (walk) which is used in its bare root form.
In addition, Recent Perfective verb forms do not assign focus, and they are not marked with the \texttt{Voice} feature.

\subsubsection{Dependency Relations}

\paragraph{Annotating clausal relations.}
In \udnewscrawl{}, we annotate the \textit{ang}-marked argument or the nominative as \texttt{nsubj}.


\begin{center}
    \begin{dependency}
        \begin{deptext}[font=\small, column sep=0.2cm]
            Nagluto \&
            ng \&
            adobo \&
            si \&
            Anna \\
            \textsc{av.prf}-cook \&
            \textsc{gen} \&
            adobo \&
            \textsc{nom} \&
            Anna \\
        \end{deptext}
        \depedge[edge height=0.4cm]{1}{5}{nsubj}
    \end{dependency}
\end{center}

An exception happens if there is an \textit{ay} inversion marker present in the clause.
In this case, the first NP is treated as the subject and the second constituent is treated as the predicate.


\begin{center}
    \begin{dependency}
        \begin{deptext}[font=\small, column sep=0.2cm]
            Ikaw \&
            ba \&
            ay \&
            magpapabakuna \&
            ? \\
            2.\textsc{sg.nom} \&
            \textsc{ques} \&
            \textsc{inv} \&
            \textsc{cau{\textasciitilde}imp-}vaccine \&
            \textsc{qm} \\
        \end{deptext}
        \depedge[edge height=1.2cm]{4}{1}{nsubj}
        \depedge[edge height=0.8cm]{4}{2}{advmod}
        \depedge[edge height=0.4cm]{3}{1}{discourse}
    \end{dependency}
\end{center}

For \textit{ng}-marked clauses, we label patient or undergoer arguments as \texttt{obj}, and label non-subject agents as \texttt{obj:agent}.


\begin{center}
    \begin{dependency}
        \begin{deptext}[font=\small, column sep=0.2cm]
            Nagluto \&
            ng \&
            adobo \&
            si \&
            Anna \\
            \textsc{av.prf}-cook \&
            \textsc{gen} \&
            adobo \&
            \textsc{nom} \&
            Anna \\
        \end{deptext}
        \depedge[edge height=0.4cm]{1}{3}{obj}
    \end{dependency}
\end{center}


\begin{center}
    \begin{dependency}
        \begin{deptext}[font=\small, column sep=0.2cm]
            Kinain \&
            ng \&
            lalaki \&
            ang \&
            adobo \\
            \textsc{<pv.prf>}eat \&
            \textsc{gen} \&
            man \&
            \textsc{nom} \&
            adobo \\
        \end{deptext}
        \depedge[edge height=0.4cm]{1}{3}{obj:agent}
    \end{dependency}
\end{center}

The UD framework only allows one argument to be labelled as the object in a sentence.
If there are cases where there are two \textit{ng}-marked arguments, we mark the more agent-like argument as \texttt{obj} and the more patient-like argument as \texttt{iobj}.


\begin{center}
    \begin{dependency}
        \begin{deptext}[font=\small]
            Pinahiran \&
            ng \&
            nanay \&
            ng \&
            Vicks \&
            ang \&
            baby \\
            \textsc{<prf>}rub\textsc{-lv} \&
            \textsc{gen} \&
            mother \&
            \textsc{gen} \&
            Vicks \&
            \textsc{nom} \&
            baby \\
        \end{deptext}
        \depedge[edge height=0.4cm]{1}{3}{obj:agent}
        \depedge[edge height=0.8cm]{1}{5}{iobj}
        \depedge[edge height=1.2cm]{1}{7}{nsubj}
    \end{dependency}
\end{center}

\paragraph{Annotating MWEs.}
Multiword expressions may be marked as having any of the three relations: \texttt{fixed}, \texttt{flat}, or \texttt{goeswith}.
We use \texttt{fixed} for fixed grammatical MWEs:


\begin{center}
    \begin{dependency}
        \begin{deptext}[font=\small, column sep=0.2cm]
            para \&
            sa \&
            mga \&
            estudyante \\
            for \&
            \textsc{dat} \&
            \textsc{pl} \&
            student \\
        \end{deptext}
        \depedge[edge height=0.4cm]{1}{2}{fixed}
        \depedge[edge height=0.8cm]{4}{1}{case}
        \depedge[edge height=0.4cm]{4}{3}{det}
    \end{dependency}
\end{center}

We use \texttt{flat} for foreign names, titles, and dates.
Finally, we use \texttt{goeswith} to link tokens that are typically joined as a single word but are separated due to the original text having typos or errors.

\subsection{Differences with UD guidelines}
Our annotation guidelines depart from the existing guidelines that were present in UD-TRG and UD-Ugnayan.
We highlight these differences below.

\paragraph*{POS Tags}
\begin{itemize}
  \setlength\itemsep{0em}
  \item Among the 17 UPOS tags, AUX was not utilized.
        Some words that express modal meaning (\textit{dapat}, \textit{puwede}) or negation (\textit{huwag}, \textit{hindi}) were marked as ADV instead.
  \item DET was only used for nominative markers \textit{ang}, \textit{si}, \textit{sina}, the plural marker \textit{mga}, and English articles \textit{the}, \textit{a}, \textit{an}.
  \item The ADP tag was used for genitive markers \textit{ng}, \textit{ni}, \textit{nina} and oblique/dative markers \textit{sa}, \textit{kay}, \textit{kina} / \textit{kila}.
\end{itemize}

\paragraph*{Morphological Features}
\begin{itemize}
  \setlength\itemsep{0em}
  \item \textbf{Nominal and Pronominal Features}
        \begin{itemize}
          \item \texttt{Gender} features were not indicated.
          \item The token \textit{sarili} is not analyzed as a pronoun. No reflexive pronouns were identified in the corpus.
        \end{itemize}
  \item \textbf{Modifier Features}
        \begin{itemize}
          \item We use \texttt{Degree=Abs} for \textit{napaka-}.
          \item We use \texttt{Degree=Equ} for \textit{kasin-/kasing}.
          \item \texttt{Polarity=Pos} was not used in the corpus.
        \end{itemize}
  \item \textbf{Verb Features}
        \begin{itemize}
          \item Two \texttt{Mood} values were used in the corpus: \texttt{Ind} (which corresponds with the \textit{realis} mood) and \texttt{Pot} (which corresponds with the \textit{irrealis} mood).
          \item Two Aspectual values were used in the corpus: \texttt{Imp} (may or may not have been initiated and not yet complete) and \texttt{Perf} (completed at a certain point of time); habitual and prospective were not used.
        \end{itemize}
  \item \textbf{Other Features}
        \begin{itemize}
          \item \texttt{Link} feature is not used in the corpus and the \textit{-ng} linker is treated as a separate token (rather than as a suffix).
          \item \texttt{NumType} feature is also indicated in the corpus.
        \end{itemize}
\end{itemize}

\paragraph*{Dependency Relations}
\begin{itemize}
  \setlength\itemsep{0em}
  \item \texttt{nsubj} subtypes were not indicated in the corpus.
  \item \texttt{csubj} was not used when a nominative marker introduces a voice-marked form to reflect the flexibility of syntactic categories in Tagalog.
  \item \texttt{case} was also used to mark the relation between a linker to a modifier.
\end{itemize}

\section{Additional description of components}
\label{appendix:components}

We train the models using the spaCy framework, resulting in a multi-task pipeline that consists of several components.
In this section, we provide additional description of these components.

\paragraph{Lemmatizer}
We use a recursive edit-tree lemmatizer based on \citet{muller-etal-2015-joint} that derives lemmatization rules from a set of examples.
For a reasonably-sized corpus, this algorithm produces thousands of edit-trees for each token-lemma pair.
In order to pick the correct edit-tree, we train a small network that learns to predict the most-probably edit-tree to lemmatize the token.
The network architecture consists of a convolutional neural network (CNN) and a layer-normalized maxout activation function.
We set the minimum frequency of an edit tree to 3, and used the surface form of the token as a backoff when no applicable edit-tree is found.

\paragraph{Morphological and POS tagger}
We employ a standard statistical approach for both morphological annotation and POS tagging by treating it as a token classification task.
A vector of tag probabilities are predicted for each token in the batch, and the most probable tag is selected for each token.
The network is then optimized using a categorical cross-entropy loss.
For the morphological analyzer, we made every unique combination of morphological features as a class.
The main limitation of this approach is that it can only predict feature combinations that exist in the training set.

\paragraph{Dependency parser}
We employ a transition-based approach to dependency parsing as described in \citet{honnibal-johnson-2015-improved}.
We have not explored graph-based algorithms such as the biaffine model from \citet{dozat-etal-2017-stanfords}, and we leave that for future work.

\section{Architecture and training hyperparameters}
\label{appendix:hyperparameters}

\subsection{Architectures}

The word-embedding models use spaCy's default token-to-vector encoding network that consists of embedding and contextual encoding subnetworks.
The hyperparameters are shown in \autoref{tab:tok2vec_hyperparams}.

{
\setlength{\tabcolsep}{2pt}
\begin{table}[h]
    \centering
    \begin{tabular}{lr}
        \toprule
        \textbf{Parameter}    & \textbf{Value} \\
        \midrule
        Dropout               & 0.1            \\
        Gradient accumulation & 1              \\
        Patience              & 1600           \\
        Max steps             & 20000          \\
        Optimizer             & Adam           \\
        $\beta_1$             & 0.9            \\
        $\beta_2$             & 0.999          \\
        L2                    & 0.01           \\
        Gradient clip         & 1.0            \\
        $\epsilon$            & 1e-7           \\
        Learning rate         & 0.001          \\
        \bottomrule
    \end{tabular}
    \caption{
        Training hyperparameters.
    }
    \label{tab:training_hyperparameters}
\end{table}
}
{
\setlength{\tabcolsep}{2pt}
\begin{table}[h]
    \centering
    \resizebox{0.95\linewidth}{!}{%
        \begin{tabular}{lr}
            \toprule
            \textbf{Parameter} & \textbf{Value}               \\
            \midrule
            \multicolumn{2}{l}{\textit{Embedding subnetwork}} \\
            Row sizes          & 5000, 1000, 2500, 2500       \\
            Attributes         & NORM, PREFIX, SUFFIX, SHAPE  \\
            \hdashline
            \multicolumn{2}{l}{\textit{Encoding subnetwork}}  \\
            Width              & 256                          \\
            Depth              & 8                            \\
            Window size        & 1                            \\
            Maxout pieces      & 3                            \\
            \bottomrule
        \end{tabular}
    }
    \caption{
        Hyperparameters of the token-to-vector encoding network.
    }
    \label{tab:tok2vec_hyperparams}
\end{table}
}

For transformer-based models that create context-sensitive vectors, we use the default parameters from HuggingFace and only set the window size for obtaining sequences for transformer processing.
For our baseline models, we set a window size of 128 and a stride of 96 characters.

\subsection{Training hyperparameters}

\autoref{tab:training_hyperparameters} shows the hyperperameters we used for training all baseline models.

\subsection{Additional results for baseline models}
\label{appendix:full_results}

Additional and fine-grained results for the morphological analyzer
and dependency parser can be found in \autoref{tab:morph_full_results} and \autoref{tab:dep_full_results} respectively.

{
\renewcommand{\arraystretch}{0.9}
\setlength{\tabcolsep}{1.2pt}
\begin{table}[h]
    \vspace{2cm}
    \centering
    \resizebox{0.95\linewidth}{!}{%
        \begin{tabular}{lcccccc}
                      & \turnbox{90}{No embeddings} & \turnbox{90}{fastText} & \turnbox{90}{Multi hash emb.} & \turnbox{90}{mDeBERTa-v3} & \turnbox{90}{RoBERTa (tl)} & \turnbox{90}{XLM-RoBERTa} \\
            \midrule
            Aspect    & 83.65                       & 88.73                  & 91.21                         & 91.54                     & 92.83                      & 92.62                     \\
            Mood      & 87.87                       & 91.47                  & 92.60                         & 93.43                     & 94.06                      & 93.86                     \\
            Voice     & 77.59                       & 79.53                  & 81.38                         & 81.62                     & 83.12                      & 82.08                     \\
            Case      & 98.88                       & 98.83                  & 98.99                         & 98.89                     & 98.88                      & 98.95                     \\
            Number    & 99.20                       & 99.18                  & 99.33                         & 99.27                     & 99.23                      & 99.34                     \\
            Person    & 99.11                       & 99.07                  & 99.31                         & 99.23                     & 99.27                      & 99.42                     \\
            PronType  & 97.69                       & 97.79                  & 98.11                         & 98.04                     & 98.21                      & 98.19                     \\
            NumType   & 91.72                       & 91.61                  & 91.75                         & 90.75                     & 91.53                      & 91.72                     \\
            Deixis    & 94.17                       & 94.17                  & 95.16                         & 94.73                     & 95.43                      & 95.25                     \\
            Abbr      & 17.45                       & 21.05                  & 2.90                          & 19.87                     & 9.79                       & 21.05                     \\
            Polarity  & 97.52                       & 97.14                  & 98.02                         & 97.90                     & 97.01                      & 98.15                     \\
            Typo      & 30.00                       & 32.79                  & 21.82                         & 44.44                     & 52.94                      & 49.32                     \\
            Degree    & 65.45                       & 65.38                  & 79.31                         & 78.69                     & 83.87                      & 90.62                     \\
            Clusivity & 98.94                       & 99.47                  & 99.47                         & 99.47                     & 98.67                      & 99.73                     \\
            PartType  & 100.00                      & 98.59                  & 92.54                         & 94.12                     & 90.91                      & 85.71                     \\
            Polite    & 87.88                       & 97.22                  & 100.00                        & 98.63                     & 100.00                     & 97.22                     \\
            \bottomrule
        \end{tabular}
    }
    \caption{
        Fine-grained morphological analysis F1-score results for all baseline models.
    }
    \label{tab:morph_full_results}
\end{table}
}

{
\renewcommand{\arraystretch}{0.9}
\setlength{\tabcolsep}{1.2pt}
\begin{table}[t]
    \vspace{2cm}
    \centering
    \resizebox{0.95\linewidth}{!}{%
        \begin{tabular}{lcccccc}
                           & \turnbox{90}{No embeddings} & \turnbox{90}{fastText} & \turnbox{90}{Multi hash emb.} & \turnbox{90}{mDeBERTa-v3} & \turnbox{90}{RoBERTa (tl)} & \turnbox{90}{XLM-RoBERTa} \\
            \midrule
            root           & 83.66                       & 84.77                  & 85.53                         & 86.60                     & 86.65                      & 88.09                     \\
            advmod         & 78.87                       & 79.37                  & 81.24                         & 81.67                     & 82.85                      & 82.52                     \\
            case           & 89.29                       & 90.07                  & 90.56                         & 90.31                     & 90.84                      & 90.87                     \\
            compound       & 27.62                       & 30.44                  & 22.61                         & 39.41                     & 38.42                      & 33.68                     \\
            obj            & 74.36                       & 74.04                  & 76.43                         & 79.80                     & 82.64                      & 81.41                     \\
            det            & 93.63                       & 93.23                  & 93.76                         & 94.09                     & 94.12                      & 94.60                     \\
            nsubj          & 80.80                       & 81.50                  & 82.31                         & 83.57                     & 85.75                      & 87.07                     \\
            nmod:poss      & 76.39                       & 78.55                  & 79.27                         & 79.10                     & 82.19                      & 81.84                     \\
            obl            & 65.50                       & 67.71                  & 68.64                         & 71.21                     & 73.55                      & 74.65                     \\
            cc             & 83.84                       & 84.68                  & 83.78                         & 87.52                     & 88.70                      & 88.30                     \\
            conj           & 61.16                       & 61.99                  & 65.87                         & 71.51                     & 71.44                      & 76.03                     \\
            fixed          & 72.71                       & 69.80                  & 72.77                         & 75.23                     & 74.80                      & 75.79                     \\
            amod           & 72.30                       & 73.22                  & 77.43                         & 75.51                     & 78.09                      & 77.89                     \\
            mark           & 78.25                       & 79.54                  & 79.37                         & 82.37                     & 83.84                      & 84.13                     \\
            acl:relcl      & 64.07                       & 64.11                  & 66.49                         & 69.01                     & 71.59                      & 75.18                     \\
            nmod           & 55.03                       & 56.96                  & 57.82                         & 60.20                     & 63.73                      & 63.72                     \\
            ccomp          & 55.39                       & 56.65                  & 57.04                         & 60.14                     & 67.91                      & 69.22                     \\
            advcl          & 49.08                       & 50.45                  & 49.72                         & 57.14                     & 61.22                      & 60.88                     \\
            flat           & 82.27                       & 82.36                  & 83.38                         & 84.48                     & 84.01                      & 86.09                     \\
            discourse      & 76.34                       & 74.62                  & 76.28                         & 79.15                     & 82.01                      & 84.01                     \\
            parataxis      & 29.69                       & 31.11                  & 30.02                         & 37.74                     & 36.88                      & 40.79                     \\
            nummod         & 80.33                       & 83.40                  & 83.94                         & 86.11                     & 85.10                      & 85.61                     \\
            obj:agent      & 78.46                       & 81.90                  & 83.24                         & 84.39                     & 87.90                      & 85.90                     \\
            dep            & 9.52                        & 14.01                  & 23.81                         & 27.23                     & 29.83                      & 22.32                     \\
            compound:redup & 47.83                       & 35.79                  & 43.90                         & 56.60                     & 56.82                      & 73.68                     \\
            xcomp          & 54.23                       & 52.86                  & 58.17                         & 58.35                     & 65.57                      & 61.47                     \\
            appos          & 53.03                       & 44.44                  & 58.29                         & 64.69                     & 60.85                      & 63.61                     \\
            acl            & 6.78                        & 9.38                   & 17.86                         & 10.67                     & 10.00                      & 21.33                     \\
            list           & 12.66                       & 16.67                  & 17.39                         & 10.17                     & 17.54                      & 17.24                     \\
            vocative       & 36.36                       & 58.33                  & 16.22                         & 71.43                     & 62.07                      & 81.48                     \\
            dislocated     & 19.18                       & 28.37                  & 31.58                         & 26.49                     & 28.36                      & 23.53                     \\
            goeswith       & 0.00                        & 0.00                   & 0.00                          & 0.00                      & 0.00                       & 0.00                      \\
            orphan         & 0.00                        & 0.00                   & 0.00                          & 0.00                      & 0.00                       & 0.00                      \\
            \bottomrule
        \end{tabular}
    }
    \caption{
        Fine-grained dependency parsing per-type LAS results for all baseline models.
    }
    \label{tab:dep_full_results}
\end{table}
}

\section{Cross-lingual performance on \udnewscrawl{}}
\label{sec:crosslingual}

\paragraph{Set-up}
In order to understand how dependency parsers trained in another language perform on the test split of \udnewscrawl{}, we follow the set-up described in \citet{aquino-de-leon-2020-parsing} and evaluate parsers trained on languages that have a sufficiently large treebank ($\geq$50k words) and are linguistically similar to Tagalog based on a metric defined by \citet{agic-2017-cross} using features from the World Atlas of Language Structures \citep[WALS,][]{Haspelmath2005WALS}.

Given a source language $S$ and target language $T$, we obtain the WALS measure $\text{dist}_W(S,T)$ by calculating the Hamming distance $d_H$ of the WALS feature vectors $\mathbf{v}_S$ and $\mathbf{v}_T$, normalized with respect to the number of features $f_{S,T}$:

$$
  \text{dist}_W(S,T) = \dfrac{d_h(\mathbf{v}_S,\mathbf{v}_T)}{f_{S,T}}
$$

For Tagalog, the most linguistically similar languages with available treebanks are Indonesian (UD-GSD), Vietnamese (UD-VTB), Romanian (UD-RRT), Catalan (UD-AnCora), and Ukrainian (UD-IU).
\autoref{tab:cross_lingual_results} shows the performance of a parser trained on each language and evaluated on the test split of \udnewscrawl{}.

{
\renewcommand{\arraystretch}{0.9}
\begin{table*}[t]
    \centering
    \begin{tabular}{@{}llrrrrrrr@{}}
        \toprule
                          &                   &                 & \textbf{Lemm.} & \textbf{UPOS} & \multicolumn{2}{c}{\textbf{Morph.}} & \multicolumn{2}{c}{\textbf{Dep. Parsing}}                               \\
        \textbf{Language} & \textbf{Treebank} & $\text{dist}_W$ & \textit{Acc}   & \textit{Acc}  & \textit{Acc}                        & \textit{F1-score}                         & \textit{UAS} & \textit{LAS} \\ \midrule
        Indonesian        & UD-GSD            & 0.446           & 65.6           & 40.3          & 50.1                                & 7.4                                       & 26.2         & 7.6          \\
        Ukrainian         & UD-IU             & 0.455           & 73.8           & 11.3          & 8.8                                 & 2.7                                       & 14.2         & 2.2          \\
        Vietnamese        & UD-VTB            & 0.469           & 62.8           & 30.6          & 12.2                                & 4.3                                       & 22.8         & 7.1          \\
        Romanian          & UD-RRT            & 0.471           & 64.6           & 34.1          & 21.8                                & 4.1                                       & 27.9         & 9.2          \\
        Catalan           & UD-AnCora         & 0.472           & 71.2           & 33.2          & 21.3                                & 5.7                                       & 25.5         & 6.8          \\ \bottomrule
    \end{tabular}
    \caption{
        Performance of a pipeline trained on a treebank of the top five languages typologically similar to Tagalog using a WALS-based metric $\text{dist}_W$ from \citet{agic-2017-cross}, as evaluated on the test split of \udnewscrawl{}.
    }
    \label{tab:cross_lingual_results}
\end{table*}
}

\paragraph{Results}
We find that there is limited cross-lingual generalization, with models trained on other languages achieving significantly lower performance on Tagalog parsing.
This suggests that Tagalog's unique syntactic patterns and typological features may require dedicated resources for effective parsing, rather than relying on transfer from other languages.
While previous studies on smaller Tagalog treebanks show more promising cross-lingual transfer \citep{aquino-de-leon-2020-parsing,miranda-2023-developing}, our results on the substantially larger \udnewscrawl{} may provide a more reliable assessment of the limitations of cross-lingual generalization.

\section{Examples for each topic in \udnewscrawl{}}
\label{appendix:topic_examples}

In this section, we illustrate some examples from \udnewscrawl{} that correspond to different topics as defined by SIB-200 \citep{adelani-etal-2024-sib}.

\begin{figure*}[t]
  \centering
  \includegraphics[width=0.8\textwidth]{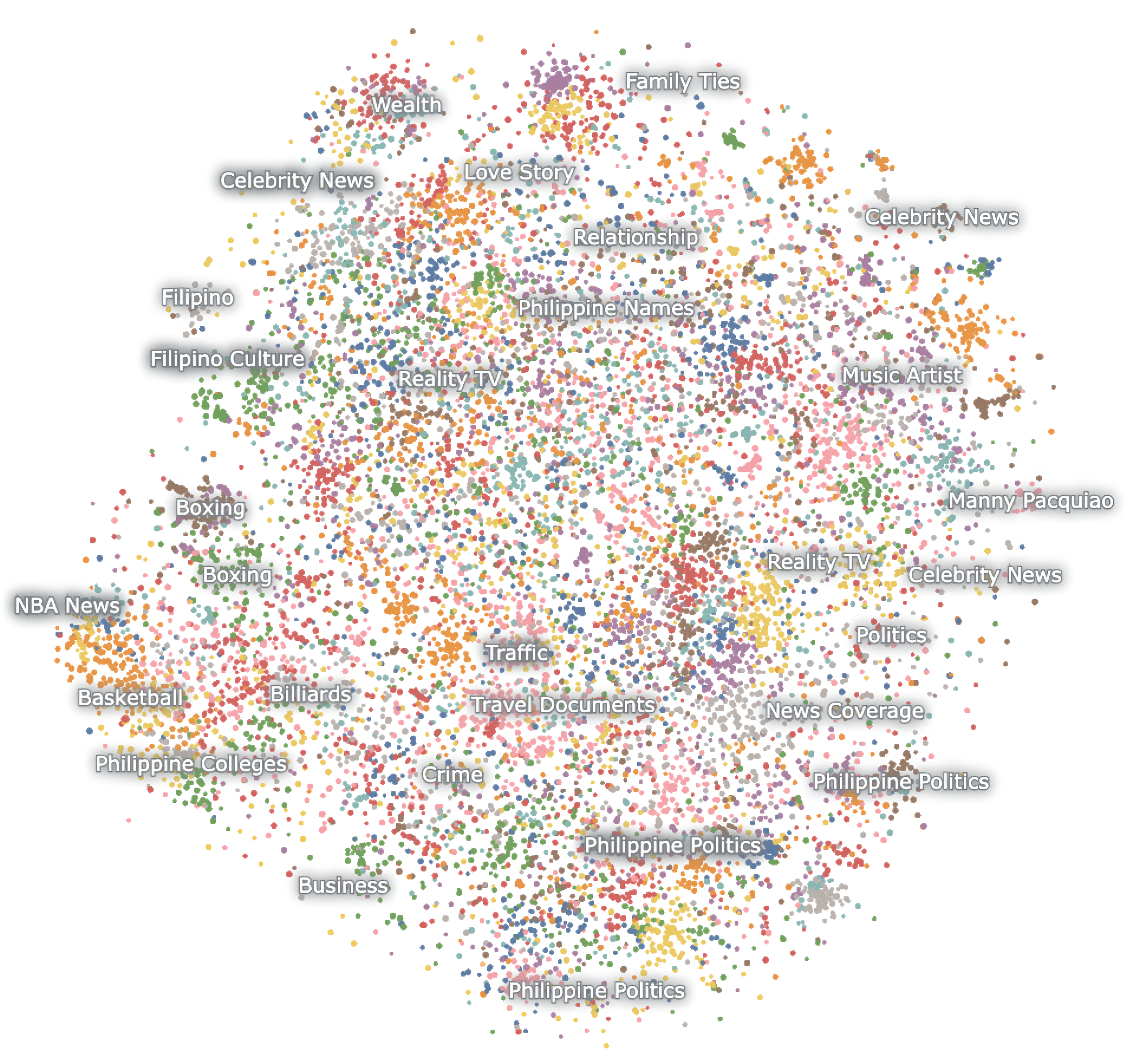}
  \caption{Embedding map generated using the Nomic Atlas API for fine-grained topic classification.}
  \label{fig:atlas_embedding_map}
\end{figure*}

\begin{itemize}
  \setlength\itemsep{-0.2em}
  \item \textbf{Entertainment}
        \begin{itemize}
          \setlength\itemsep{-0.2em}
          \item \textit{Alam din daw niyang masaya na si Nikki sa boyfriend nitong si Billy Crawford.} (He said he also knows that Nikki is already happy with her boyfriend Billy Crawford.)
          \item \textit{Out of the blue, sumali kami sa kaswal na chikahan nina Allan at direk Dom.} (Out of the blue, we joined in the casual chat between Allan and director Dom.)
          \item \textit{Sa issues na tatalakayin sa show, aamin na kaya si Bianca King kung siya na ang bagong Reyna ng Kasungitan na ibinabato sa kanya?} (Among the issues that will be discussed on the show, will Bianca King finally admit if she is the new Queen of Anger that is being thrown at her?)
        \end{itemize}
  \item \textbf{Health}
        \begin{itemize}
          \setlength\itemsep{-0.2em}
          \item \textit{Kailangan umanong masuportahan ito ng dokumento o ebidensya, tulad ng resulta ng drug test.} (This must be supported by documents or evidence, such as drug test results.)
          \item \textit{Kasi tumaba na ako, ang hirap kayang magpapayat.} (Because I've gained weight, it's hard to lose weight.)
          \item \textit{Noong 1983 pa nagkaroon na ng product recall ang Saridon matapos mapatunayan ng US FDA na maaari itong pagmulan ng cancer sa mga taong umiinom ng mga gamot na may phenacetin (kasama na ang Saridon).} (Saridon had a product recall as early as 1983 after the US FDA confirmed that it could cause cancer in people taking medications containing phenacetin (including Saridon).)
        \end{itemize}
  \item \textbf{Sports}
        \begin{itemize}
          \setlength\itemsep{-0.2em}
          \item \textit{Ititiklop ng Elasto Painters ang eliminations kontra Alaska bukas sa Ynares Center sa Antipolo.} (The Elasto Painters will repeat the eliminations against Alaska tomorrow at the Ynares Center in Antipolo.)
          \item \textit{Simula na rin ng banggaan ng defending NBA champion at Eastern Conference No. 2 seed Boston laban sa 7th seeded Chicago Bulls.} (The clash between defending NBA champion and Eastern Conference No. 2 seed Boston and the 7th seeded Chicago Bulls has begun.)
          \item \textit{Dalawang larong liban si Taulava sa pagsaklolo sa Smart Gilas Pilipinas na pumang-anim nga lang sa katatapos na Guangzhou Asian Games.} (Taulava missed two games to help Smart Gilas Pilipinas, which only finished sixth in the recently concluded Guangzhou Asian Games.)
        \end{itemize}
  \item \textbf{Politics}
        \begin{itemize}
          \setlength\itemsep{-0.2em}
          \item \textit{Ang tanging kalaban ni Arroyo sa nasabing posisyon ay ang private citizen na si Adonis Simpao.} (Arroyo's only opponent in the said position is private citizen Adonis Simpao.)
          \item \textit{Samantala, mabilis na itinanggi ni Sen. Jinggoy ang isyung ang kampo nila ang nagpakalat na bayad ang pag-eendorso ni Dolphy kay presidential aspirant Manny Villar.} (Meanwhile, Sen. Jinggoy quickly denied the issue that his camp spread the rumor that Dolphy's endorsement of presidential aspirant Manny Villar was paid for.)
          \item \textit{Ang kapatid ngayon ni Nograles na si Jose ang presidente ng PDIC.} (Nograles' brother Jose is now the president of PDIC.)
        \end{itemize}
  \item \textbf{Travel}
        \begin{itemize}
          \setlength\itemsep{-0.2em}
          \item \textit{Dumating kami dito sa Gold Coast noong Martes ng umaga mula sa mahabang biyahe mula sa Maynila.} (We arrived here on the Gold Coast on Tuesday morning after a long trip from Manila.)
          \item \textit{Ang sabi sa amin, ang biyahe mula Maynila hanggang Santa Ana Park ay inaasahang iiksi ng isang oras na lamang kapag nagawa at napadaanan na ang nasabing highway.} (We are told that the trip from Manila to Santa Ana Park is expected to be shortened to just one hour once the said highway is completed and passable.)
          \item \textit{Hindi gaanong malalaki ang mga gusali at limitado hanggang alas-siyete ng gabi ang operasyon ng mall nila doon.} (The buildings are not very large and their mall's operations are limited to seven o'clock in the evening.)
        \end{itemize}
  \item \textbf{Geography}
        \begin{itemize}
          \setlength\itemsep{-0.2em}
          \item \textit{Malaysia, Singapore, at ibang karatig-bansa.} (Malaysia, Singapore, and other neighboring countries.)
          \item \textit{Ang mga puno sa magkabilang hanay ng MacArthur Highway mula City of San Fernando hanggang sa Angeles City ay nagbibigay ng sariwa at luntiang damdamin sa sinumang dumadaan dito.} (The trees on both sides of the MacArthur Highway from the City of San Fernando to Angeles City give a fresh and green feeling to anyone passing by.)
          \item \textit{Agrikultura ang major industry ng lugar kaya’t milya-milya ng lupaing tinamnan sa palay, maize, niyog at marami pang iba ang makikita ng bisitang nagbabaybay ng probinsya.} (Agriculture is the area's major industry, so visitors traveling through the province will see miles and miles of land planted with rice, maize, coconuts, and many other crops.)
        \end{itemize}
  \item \textbf{Science/Technology}
        \begin{itemize}
          \setlength\itemsep{-0.2em}
          \item \textit{Ginawan ng maraming pag-aaral tulad ng pagsasagawa ng Parañaque Spillway na bibigyang-daan ang daloy ng tubig galing sa Laguna Lake papuntang Manila Bay.} (Many studies have been conducted, such as the construction of the Parañaque Spillway, which will allow the flow of water from Laguna Lake to Manila Bay.)
          \item \textit{Ang PRES ay isang electric service na gagamit ng prepaid metering system na layong payagan ang mga residential customers na bumili ng credit o load para sa paggamit ng kuryente hanggang sa ito'y maubos.} (PRES is an electric service that will use a prepaid metering system that aims to allow residential customers to purchase credit or load for electricity use until it is exhausted.)
          \item \textit{Ayon kay Philippine Atmospheric, Geophysical and Astronomical Services Administration (PAGASA) Director Prisco Nilo, karaniwang nagaganap ang meteor shower tuwing buwan ng Agosto.} (According to Philippine Atmospheric, Geophysical and Astronomical Services Administration (PAGASA) Director Prisco Nilo, meteor showers usually occur every August.)
        \end{itemize}
\end{itemize}

\section{Fine-grained topic classification}
\label{appendix:fine_grained_topic_clf}

We also perform fine-grained topic classification using a multilingual generate text embedding (GTE) model \citep{zhang2024mgte}.
We then use the Nomic Atlas API\footnote{https://atlas.nomic.ai/} to perform topic modelling using a hierarchical clustering model.
The embedding map can be found in \autoref{fig:atlas_embedding_map}.

The embedding map suggests that \udnewscrawl{} captures multiple domains of Philippine society, with 20 major topic clusters identified through our analysis.
These clusters encompass several key categories: entertainment (Celebrity News, Reality TV, Love Story), sports (Basketball, NBA News, Boxing), politics (Philippine Politics appearing in multiple distinct clusters), and cultural elements (Filipino Culture, Philippine Names).

Our findings suggest that \udnewscrawl{} captures a diverse range of topics in Philippine news media, with particularly strong representation of entertainment, sports, and political content.
This topical distribution, as revealed through the embeddings, is consistent with our earlier topic analysis findings in \S\ref{sec:topic_clf}, reflecting the typical content priorities and coverage patterns of mainstream Philippine news outlets.

\section{Prompt template for topic classification}
\label{appendix:prompt_topic}

\begin{figure*}[h]
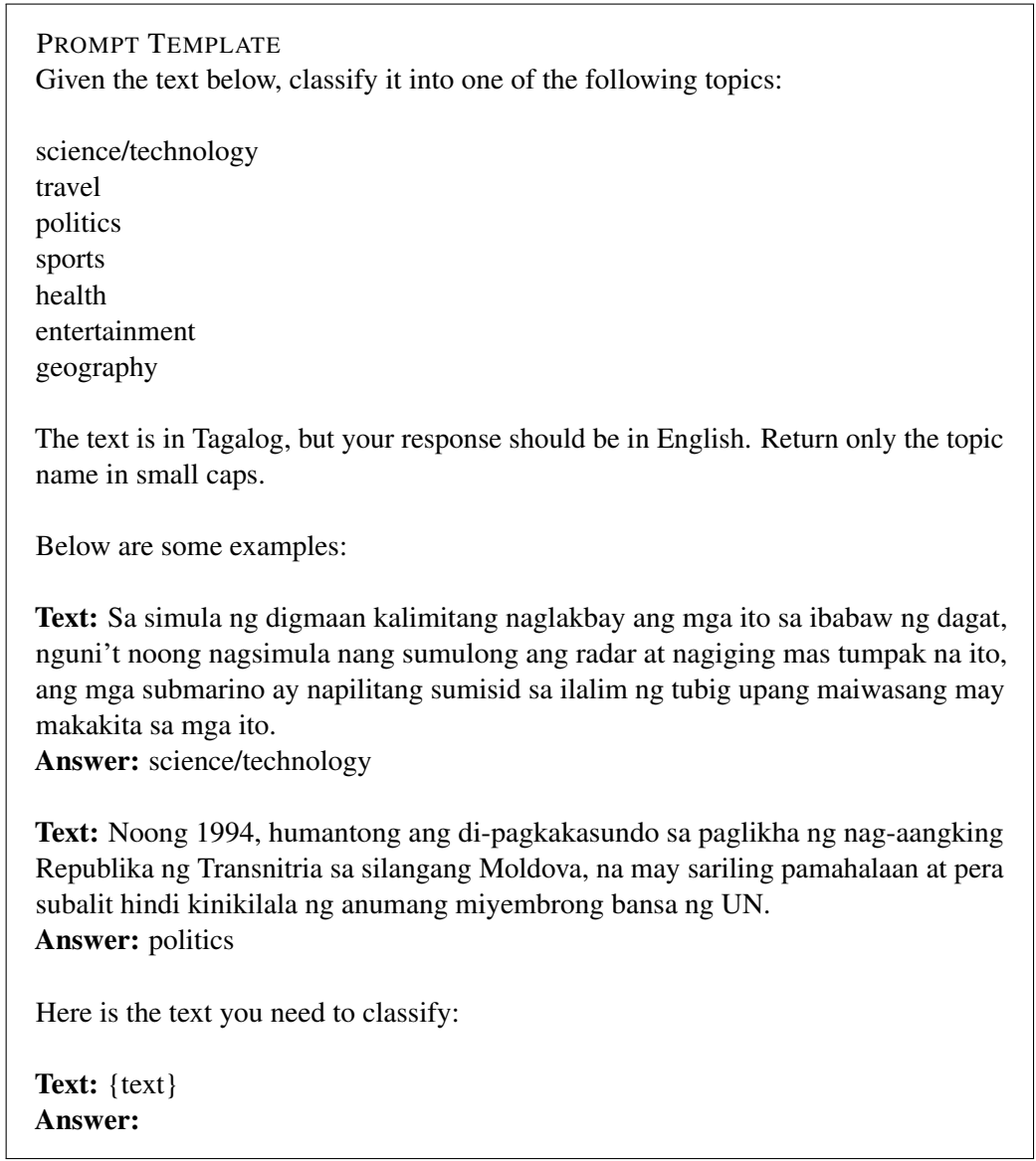

    \centering
  {
    \setlength{\fboxsep}{1em}
    \noindent\fbox{%
      \parbox{0.80\linewidth}{%
        \textsc{Prompt Template}\\
        Given the text below, classify it into one of the following topics:\\

        science/technology\\
        travel\\
        politics\\
        sports\\
        health\\
        entertainment\\
        geography\\

        The text is in Tagalog, but your response should be in English.
        Return only the topic name in small caps.\\

        Below are some examples:\\

        \textbf{Text:} Sa simula ng digmaan kalimitang naglakbay ang mga ito sa ibabaw ng dagat, nguni't noong nagsimula nang sumulong ang radar at nagiging mas tumpak na ito, ang mga submarino ay napilitang sumisid sa ilalim ng tubig upang maiwasang may makakita sa mga ito.\\
        \textbf{Answer:} science/technology\\

        \textbf{Text:} Noong 1994, humantong ang di-pagkakasundo sa paglikha ng nag-aangking Republika ng Transnitria sa silangang Moldova, na may sariling pamahalaan at pera subalit hindi kinikilala ng anumang miyembrong bansa ng UN.\\
        \textbf{Answer:} politics\\

        Here is the text you need to classify:\\

        \textbf{Text:} \{text\}\\
        \textbf{Answer:}
      }%
    }
  }
  \caption{Prompt template for topic classification.}
  \label{fig:prompt_template}
\end{figure*}

We provide the prompt template for the topic classification experiment in \autoref{fig:prompt_template}.
We find that writing the instructions in English (instead of Tagalog) and specifying the language of the text-in-question (``The text is in Tagalog...'') gives more consistent and parseable outputs in our evaluation.

\section{Error analysis}
\label{appendix:error_analysis}

During quality control (\S\ref{sec:qualitycontrol}), we compare our expert annotations with the existing global and language-specific rules in the UD framework using the UD validator.
We find that there are two major error categories in our expert annotations: incompatibilities between particular UPOS tags (L3) and language-specific labels that do not yet exist in the UD guidelines (L4).\footnote{\url{https://universaldependencies.org/validation-rules.html}}

\begin{figure}[t]
  \centering
  \includegraphics[width=\linewidth]{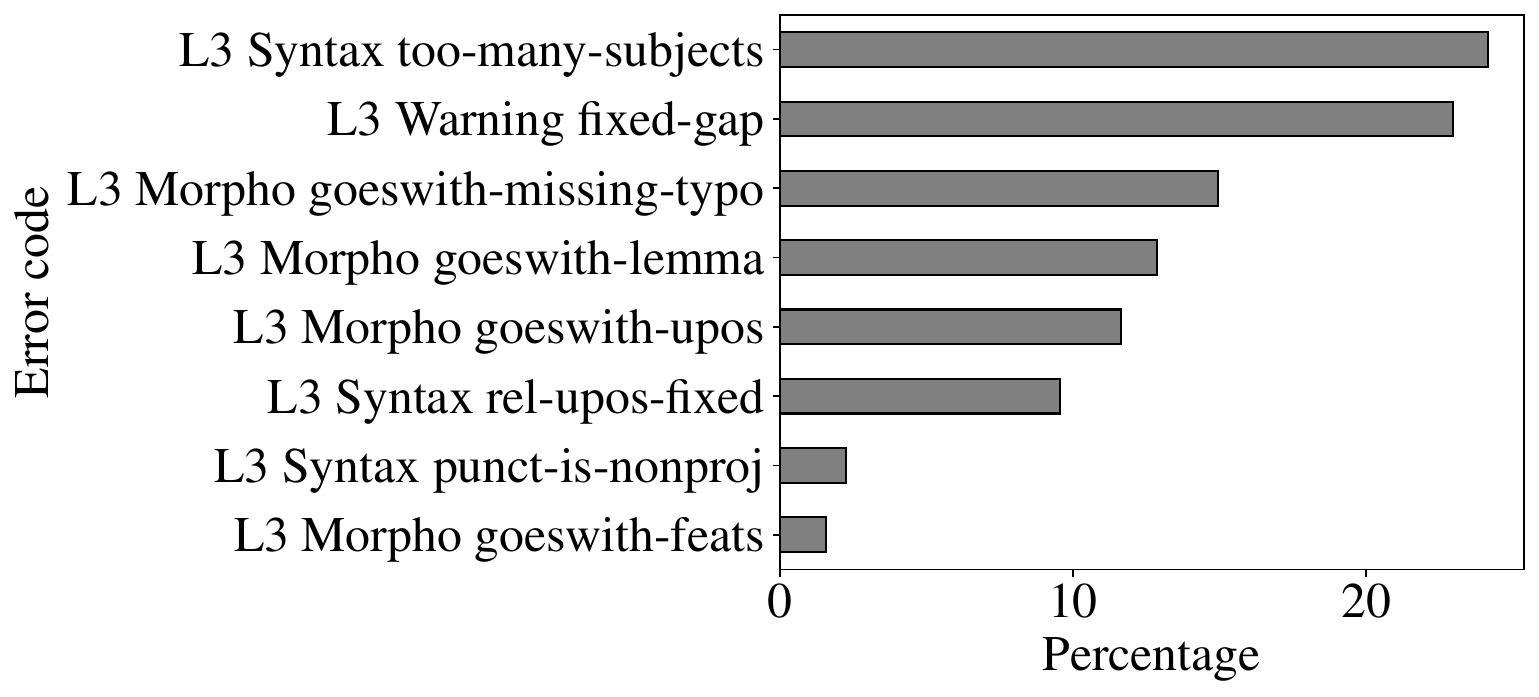}
  \caption{Breakdown of L3 errors that relate to incompatibilities between particular UPOS tags as detected by the official UD validator.}
  \label{fig:l3_errors}
\end{figure}

\subsection{Type L3 Errors}
\autoref{fig:l3_errors} show a breakdown of L3 errors.
We find that some of these are due to the annotation decisions we made regarding Tagalog grammar as described in \S\ref{sec:challenges} (and Appendix \ref{appendix:annotation_guidelines}) which contradict some aspects of the universal guidelines.

\subsection{Type L4 Errors}

The majority of L4 errors include language-specific labels that affect morphological features and dependency relations.
For example, the validator might say that a morphological feature is not permitted with a certain UPOS (e.g., ``Feature Case is not permitted with UPOS DET in language [tl]'').
Resolving these errors involve either one or two fixes: update the language-specific guidelines to reflect new cases found in our treebank or correct the expert annotations.
We find that most of the fixes involve the former approach, as the language-specific guidelines for Tagalog are still sparse.

This pattern is entirely expected, given that the existing Tagalog treebanks are relatively small in size and scope.
The current language-specific guidelines were likely derived from these limited samples, which may not capture the full range of linguistic phenomena present in Tagalog.
Our larger and more diverse treebank effectively serves as a window into previously undocumented morphosyntactic patterns, revealing cases where determiners, for instance, can indeed carry case marking in specific contexts.
This presents a valuable opportunity to expand and refine the language-specific guidelines, making them more comprehensive and representative of actual language usage. Rather than viewing these L4 errors as mere validation issues, they can be seen as important signals highlighting areas where our understanding of Tagalog's formal grammatical structures needs to be updated and codified in the Universal Dependencies framework.

\end{document}